# SEOpinion: Summarization and Exploration Opinion of E-Commerce Websites

Alhassan Mabrouk [1], Rebeca P. Díaz Redondo [2], Mohammed Kayed [3,*]

[1] Mathematics and Computer Science Department, Faculty of Science, Beni-Suef University, Egypt; h.mabrouk100@gmail.com
[2] Information & Computing Lab, AtlantTIC Research Center, Telecommunication Engineering School. Universidade de Vigo, Spain; rebeca@det.uvigo.es
[3] Computer Science Department, Faculty of Computers and Artificial Intelligence, Beni-Suef University, Egypt
* Correspondence: mskayed@gmail.com

**Abstract:** E-Commerce (EC) websites provide a large amount of useful information that exceed human cognitive processing ability. In order to help customers in comparing alternatives when buying a product, previous studies designed opinion summarization systems based on customer reviews. They ignored templates' information provided by manufacturers, although these descriptive information have much product aspects/characteristics. Therefore, this paper proposes a methodology coined as SEOpinion (Summarization and Exploration of Opinions) which provides a summary for the product aspects and spots opinion(s) regarding them, using a combination of templates' information with the customer reviews in two main phases. First, the Hierarchical Aspect Extraction (HAE) phase creates a hierarchy of product aspects from the template. Subsequently, the Hierarchical Aspect-based Opinion Summarization (HAOS) phase enriches this hierarchy with customers' opinions; to be shown to other potential buyers. To test the feasibility of using Deep Learning-based BERT techniques with our approach, we have created a corpus by gathering information from the top five EC websites for laptops. The experimental results show that Recurrent Neural Network (RNN) achieves better results (77.4% and 82.6% in terms of F1-measure for the first and second phase) than the Convolutional Neural Network (CNN) and the Support Vector Machine (SVM) technique.

## 1. Introduction

E-Commerce (EC) websites support a quick expansion of consumer reviews and feedback (e.g., Amazon, eBay, Flipkart and Snapdeal). These reviews are valuable to both product developers (marketers) and consumers. Product developers are interested in identifying those aspects (attributes) that are important for consumers [1]. Besides, consumers arguably make purchase decisions based on previous evaluations from other customers [2]. With the development of EC websites, popular products (e.g., iPhone and Samsung) contain large amounts of review text. So, summarizing the aspects of these products by humans is a complex and time-consuming process [3]. Thus, it is essential to provide an Aspect-based Opinion Summarization (AOS) [4]. This paper improves AOS using Deep Learning (DL)-based BERT embedding [5]. Recently, DL technologies have been applied with big data analytics. Everywhere, big data are part of many information processing systems like science, government, health care [6], security/privacy, financial [7] and social media [8], [9], [10].

AOS is a sentiment analysis task that can summarize opinions on aspects given a set of reviews. This task always involves three phases [11]: (i) aspect extraction [12] (ii) aspect-level polarity detection and (iii) summary presentation. First, the aspect extraction phase fetches the topics from the review text [13]. For example, a sentence such as "The screen of my laptop is nice and its resolution is good" has two aspects, namely, the "screen" and the "resolution". Second, the aspect-level polarity detection phase determines the sentiment orientation (positive or negative) on the extracted aspects. In the above example, the sentence has two positive aspects: "screen" and "resolution". The two previous phases of automatic aspects extraction and polarity/strength prediction are jointly called aspect-based sentiment analysis in which more information is provided [14]. Finally, in the summary presentation phase, the processed results are presented by aggregating polarity ratings for all aspects and summarizing opinions around them.

The extracted aspects in AOS systems are represented using two different structures: a flat structure and a hierarchical structure. Using the flat structure means the aspects of a specific domain are represented as a list [15]. For example, a laptop is represented as a list of the two aspects "screen" and "resolution" in the example above. On the other hand, using the hierarchical structure means the aspects of a specific domain are structured into multi-granularity of aspects [16]. For example, the hierarchical structure of the same example has two levels in addition to the root ("laptop"), in which the "screen" (level 1) is the aspect of a laptop while the "resolution" (level 2) is the aspect of the "screen". Most approaches of AOS ignored the hierarchical structure inside the aspects [17], [18] . Although, it is much valuable to buyers and product manufacturers in understanding the accurate aspects quickly inside the massive consumer reviews, in addition to the hierarchical nature of aspect terms. Alternatively, few researchers have targeted to summarize opinions of multi-granular aspects, which seems to be more appropriate than flat aggregation [19].

The problem of hierarchical aspects-based opinion summarization is a challenge [19], such as aspects hierarchy is provided manually (i.e., a predefined structure) [16], [20], or a large amount of training data are needed for a summary presentation [21]. These problems entail the current hierarchical AOS approaches are not-scalable. So, this paper proposes an automated approach called SEOpinion that extracts popular aspects from the product details (i.e., the templates of the websites) in a hierarchical structure and classifies the opinionated sentences (from customer reviews) according to their aspects. Our proposed approach includes five main tasks (see Figure 1). In the first task, the aspects are extracted from a set of product details (e.g., HP, Dell, and Apple) of the same product type (i.e., Laptop), as shown in Figure 1 (a). The second task constructs an aspect hierarchy using the extracted aspects. The third task extracts opinionated sentences from the product reviews. The fourth task automatically maps the aspects in the hierarchy to extracted opinion sentences. For instance, the opinionated sentence "This laptop is ok for its price" matches the aspect "price". The fifth task classifies sentiment polarity (positive or negative) of each opinionated sentence associated with its aspect to be ready for the summarization.

The advantages of our SEOpinion system are: (i) it helps users to easily access a sentiment score about any preferred aspect and all sub-aspects thereof; (ii) the extracted aspects do not depend on a specific domain (e.g., Amazon, Flipkart, or others), as long as these aspects are directly extracted from the site templates; (iii) the extracted hierarchy of aspects is constant and does not change with the change of reviews as in other methods [22], [23], because they have been obtained from product details, not from the reviews (as shown in step 1 of Figure 1); (iv) some users might prefer to read the actual opinion sentences instead of reading the overall statistics, so these are displayed in a separate panel, called the opinion sentence exploration (the details are shown in Section III.E); (v) it allows users to easily compare people's opinions on products of the same type (e.g., camera) because they are all represented by the same aspects (i.e., Zoom, Lens, Focus and etc.); and finally (vi) it also helps in the polarity classification processes, which show the polarities of some sentiment words. For example, the sentence "In this laptop, the processor and battery-life are fast" contains a sentiment word "fast" and two aspects "processor" and "battery-life". The "fast" is positive under the "processor" aspect node, while it is negative under "battery-life".

To the best of our knowledge, no prior works have focused on summarizing opinions of hierarchical aspects extracted from product details. Thus, our main contributions are as follows:

1. Create a web scraper to crawl the product details and reviews from e-commerce websites using XPath (XML path language);
2. Hierarchical representation of the relevant product aspects, which are obtained from the product details and descriptions published in the web pages by the manufacturers;
3. Each review sentence is mapped directly to its corresponding aspect;
4. For each product aspect, the sentiment-score and opinionated sentences are shown;
5. Create a corpus to validate the proposed approach, which is obtained from the top five EC (laptops) websites;
6. Our results show that the usage of BERT [5] embedding on RNN model gives better results than CNN and SVM on our corpus.

The rest of the paper is organized as follows. Section 2 presents related works about aspect-based opinion summarization. The proposed system is discussed in Section 3. Section 4 shows the details of the experiment. The results of our experiment and their analysis are given in section 5. A discussion of the limitations of our proposed system, and the future directions are detailed in Section 6. Finally, Section 7 concludes our work.

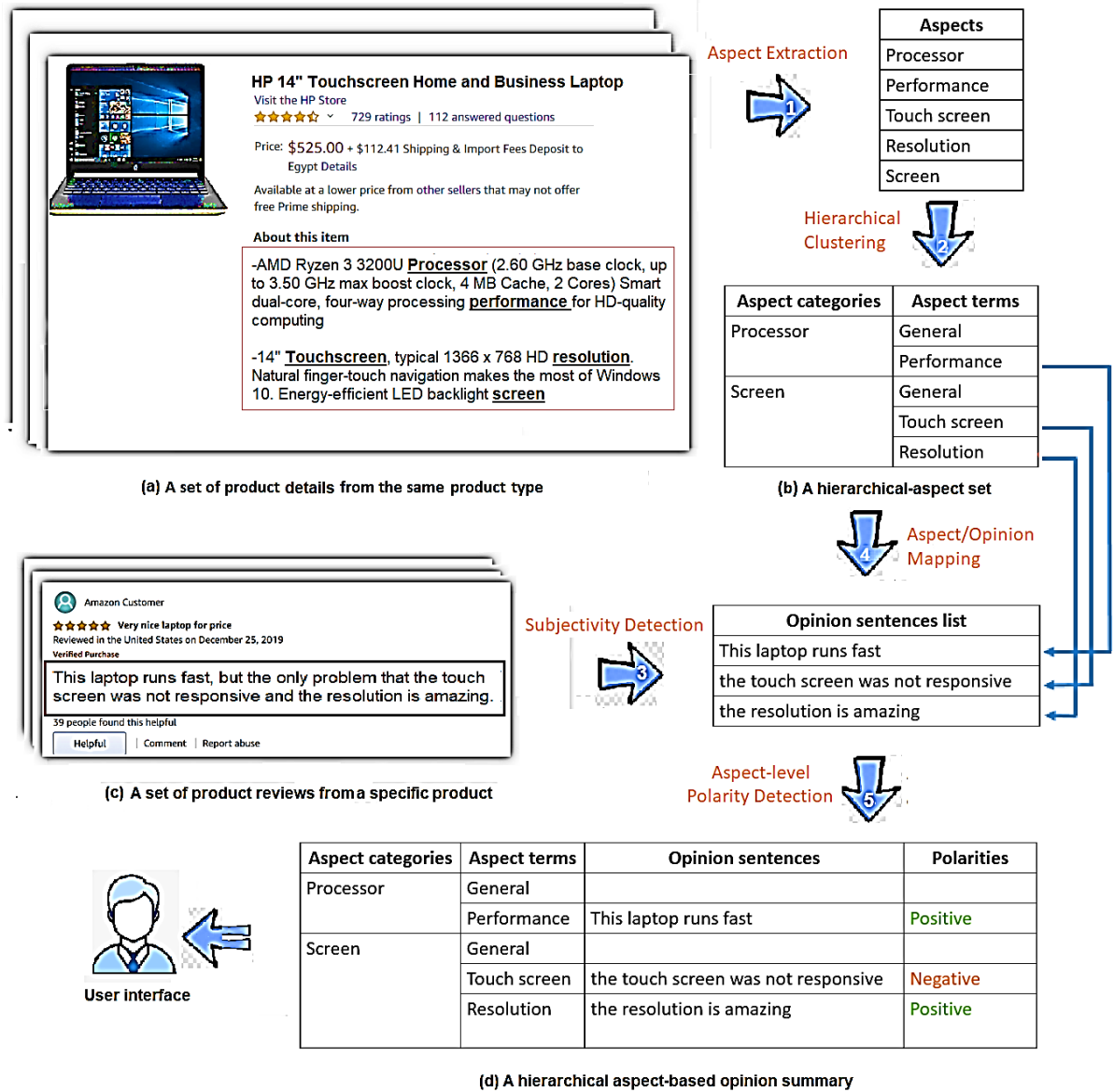

**FIGURE 1.** The steps of the proposed approach: SEOpinion.

## 2. Related Works

*2.1. Aspect-Based Opinion Summarization*

Most existing studies have been devoted to performing the Aspect-based Opinion Summarization (AOS) task [15], [17], [18]. Bahrainian and Dengel [17] used a hybrid polarity detection method in order to summarize aspects of multiple documents. They used topic detection algorithms to discover different domain-based lexicons. Their algorithm recognized newly added features, but less accurate than manual detection. Zhu et al. [15] suggested a framework for summarizing the opinion based on the sentence. Besides, they expected the review's helpfulness, taking into account coverage and frequency. Jmal and Faiz [18] introduced the aspect summary approach for the entire product by calculating a score between 0-1 for its characteristics based on adverbs, nouns, verbs, and adjectives. However, these mentioned methods extracted their aspects in a flat structure without considering the natural hierarchical structure inside the aspects.

Few methods addressed the hierarchical structure of the aspects to be extracted [24]. In sentiment analysis, Multi-grain Topic Model [25] handled multiple granularities of aspects in the context of online review analysis. Kim *et al.* [26] proposed an unsupervised the Hierarchical Aspect Sentiment Model (HASM) to discover a hierarchical structure of aspect summarization from unlabeled online reviews. HASM deals aspect extraction with sentiment modeling. Almars *et al.* [27] proposed hierarchically model users' interests and sentiments on various topic levels in the tree. However, these models were proposed for aspect identification and their effectiveness were not investigated for sentiment summarization.

*2.2. Hierarchical Opinion Summarization*

Some researchers have targeted summarizing opinions of multi-granular aspects, which are more appropriate than flat aggregation, due to the hierarchical nature of the aspect terms. Yu et al. [16] proposed a domain-assisted approach for reordering product aspects in a hierarchical structure. First, they automatically obtained a primary hierarchy, then detected product features from the reviews. And finally, the semantic distance between aspects helped for inserting them into the initial hierarchy. Pavlopoulos and Androutsopoulos [19] introduced a domain-independent method to group the aspects. They also investigated word-vectors based on WordNet to calculate the similarities between words in the hierarchical clustering algorithm. However, some problems have been identified in the existing hierarchical aspects of summarization systems, as ontology tree generated manually, domain-specific, or pre-defined.

Surprisingly, recent research work on hierarchical aspect aggregation [22] proposed an automatic approach to generate an aspect ontology tree using similarity techniques that work across domains. They did consider WordNet in aspect aggregation, although word embedding can overcome such limitations encountered in WordNet. In contrast, in [28], they built the hierarchy using word embedding to represent each aspect by a vector and then clustering those vectors. However, these mentioned methods are distinct taxonomies that can be generated for two products of the same type. On the other hand, OpinionLink enriched the product aspects with opinions extracted from user reviews, where these aspects have been designed via human readers, as proposed in [23]. In contrast, our system automatically discovers product aspects using information from the webpage templates.

In order to enhance the results of the aspect extraction process, some approaches exploited the product reviews as well as terms embedded in the Webpage template of the product [29]. They used the structure list of the template for extracting the aspects [29]. However, getting structured product specifications are very expensive. In contrast, in [30], they incorporated customer reviews and product descriptions, which are cheap because manufacturers almost provide this information for users. However, to the best of our knowledge, these methods did not address/show the effect of applying these extracted aspects from page' templates on the opinion summarization problem. Therefore, this paper addresses the problem of mapping opinionated sentences with the extracted hierarchical-aspects from templates.

**3. SEOpinion: Methodology**

This section describes the proposed Summarization and Exploration Opinion (SEOpinion) system in detail. The first subsection gives an overview of the whole system architecture. Afterward, the scraping process, hierarchical aspect extraction (Phase A) and hierarchical aspect-based opinion summarization (Phase B) are addressed, respectively. Lastly, the SEOpinion's interface is discussed.

*3.1. Overview*

The general structure of our SEOpinion system is given in Figure 2. The system takes a set of product web page templates of the same product type as input and generates a set of summaries for these products as output, which each template has product details and customer reviews. The SEOpinion system is split into web scraping and two other main phases: Hierarchical Aspect Extraction (HAE) and Hierarchical Aspect-based Opinion Summarization (HAOS). In the web-scraping phase, the product details and reviews are crawled from EC websites. After that, in the HAE phase, the popular product aspects are extracted from the product details of the same type, and then stored in a hierarchical format, as shown in Figure 1 (b). In the HAOS phase, opinions are first extracted from reviews for each product separately. After that, these opinions are mapped to the hierarchical-aspect set. Furthermore, the polarity of each one is classified (i.e., positive or negative) according to the aspect associated with it, as shown in Figure 1 (d).

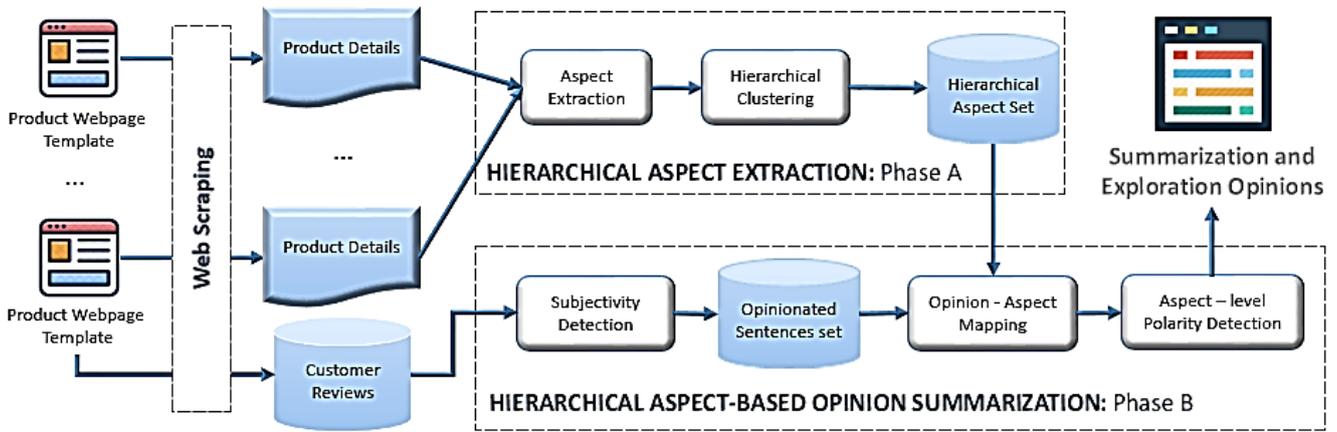

**FIGURE 2.** Overview of the SEOpinion system.

The Proposed Methodology Is Described Using The Pseudocode In Algorithm 1. The Input Is A Set Of Product Web Page Templates Of The Same Type P = {P1, P2 … Pn}, While The Output Is A Summarization Of Aspects And An Exploration Of The Gathered Opinions. Step 1 Initializes Our System. Step 2 Scraps All Product Details D And Reviews R From Product Templates P. The Scraping Process Is Encapsulated In The Scraping Function (Step 2). The Function *HAExtraction* (Step 3) In The Algorithm Creates A Hierarchy Of Common Aspects H From All Product Details In P. After That, The Algorithm Iterates Through All Products In P (Loop 4 - 8). In Each Iteration (I.E., On Each Product Separately), The Function *HAOSummary* (Step 6) Produces An Aspect Summary Si Based On The Opinions That Match It. Finally, Summarization Aspects And Exploration Opinionated Sentences Are Obtained.

**Input:**

   *P:* A set of products web page templates of the same product type = $\{p_1, p_2 \dots p_n\}$

**Output:**

   *SEO:* Summarization and exploration opinions

**Method:**

*1:* $SEO \leftarrow \phi$

*2:* $R, D \leftarrow Scraping\ (P)$

*3:* $H \leftarrow HAExtraction\ (D)$                                    ⇒ *Phase A*

*4:* **for each** product $p_i \in P$ **do**

*5:*        let $R_i \in R$, be a set of reviews in $p_i$

*6:*        $S_i \leftarrow HAOSummary\ (R_i, H)$                   ⇒ *Phase B*

*7:*        $SEO \leftarrow SEO \cup S_i$

*8:* **end for**

*9:* **return** *SEO*

**ALGORITHM 1.** SEOpinion system.

*3.2. Web Scrapping*

    Web scraping is a technique used to fetch data from websites using web scrapers/crawlers. Web scrapers are scripts that use the HTTP (Hypertext Transfer Protocol) to connect to the World Wide Web (WWW) and allow users to retrieve information. Therefore, our web scraper has been built to collect product information from the EC websites (there are

TABLE 1. Samples of different structures in the top five E-commerce websites.

| EC website | Type | Useful data parts | XPath Format |
|---|---|---|---|
| Amazon | Product details | - Title | //span[@id='productTitle']/text() |
| | | - About this item* | //div[@id='feature-bullets']/ul/li/span/text() |
| | | - Compare with similar items | //table[@id='HLCXComparisonTable']//tr/th/span/text() |
| | | - Product description* | //div[@id='productDescription']/text() |
| | | | //div[@id='productDescription']/b/text() |
| | | - Product information | //table[@id='productDetails_techSpec_section_1']//tr/th/text() |
| | | | //table[@id='productDetails_techSpec_section_2']//tr/th/text() |
| | Customer Reviews | | //div[@data-hook='review-collapsed']/span/text() |
| Flipkart | Product details | - Title | //span[@class='_35KyD6']/text() |
| | | - Highlights* | //div[@class='_3WHvuP']/ul/li/text() |
| | | - Description* | //div[@class='_3la3Fn _1zZOAc']/p/text() |
| | | - Specifications | //table[@class='_3ENrHu']/tbody/tr/td[1]/text() |
| | Customer Reviews | | //div[@class='qwjRop']/div/div/text() |
| eBay | Product details | - Title | //h1[@id='itemTitle']/text() |
| | | - Item specifics | //td[@class='attrLabels']/text() |
| | | - About this product | //div[@class='prodDetailSec']/table/tbody/tr/td[1]/text() |
| | | -Review Text | //div[@class='ebay-review-section-r']/p/text() |
| | Customer Reviews | | //div[@class='ebay-review-section-r']/p/text() |
| Walmart | Product details | - Title | //h1[@itemprop='name']/text() |
| | | - About This Item* | //div[@class='about-desc about-product-description xs-margin-top']/text() |
| | | | //div[@class='about-desc about-product-description xs-margin-top']/ul/li/text() |
| | | - Specifications | //table[@class='product-specification-table table-striped']/tbody/tr/td[1]/text() |
| | Customer Reviews | | //div[@class='review-text']/p/text() |
| BestBuy | Product details | - Title | //h1[@itemprop='name']/text() |
| | | - Other Specifications | //table[@class='product-spec']/tr/th/text() |
| | | | //table[@class='product-spec']/tr/td[1]/text() |
| | | - Description* | //div[@itemprop='description']/text() |
| | Customer Reviews | | //div[@class='user-review']/p/text() |

two types of information in these sites, namely, product details and customer reviews). On creating the scraper, we faced three main challenges. First, generalizing websites is challenging because the templates vary from site to site. Second, web page structures are constantly updated by web developers. Hence, it is difficult to rely on a single scraper for a long time. Third, the structure of the same website may differ from one category to another. For example, on the Amazon website, the structure of the Electronic category is different from the structure of the Book category in the template format. To address the above challenges, we follow the continuous development and integration of a specific domain on EC sites using the XPath (XML Path Language) query language. XPath contains the path of any element located on the web page, which is simple, powerful, concise, and easy to get. Table 1 shows the recent structure update of the extracted parts from the Laptop domain on the top five EC websites using the XPath format.

This paper uses the Scrapy[1] python package as in [31], [32] to create the web scraper. Scrapy was designed to scrape the web content from websites that are composed of many pages of similar semantic structure (i.e., the templates of web pages). Scrapy stands out from any other scraping tools (e.g., Selenium and Apify) because it is faster, parallel processing and can deal with structured data and open-sources. The steps to create our scraper are summarized as follows. First, the web scraper visits publicly available web pages that contain product details and reviews. Then, it receives HTML (Hypertext Markup Language) data back from the web server, in which the content of the web pages is embedded. After that, the Scrapy extracts the useful-data parts from the HTML using the XPath format. The code snippet "response. xpath ('XPath Format'). getall ()" is used to scrape data from several EC websites using Table 1 to know the XPath format structure on the top five EC websites. Finally, our scraper stores these data in a JSON file (JavaScript Object Notation). Each JSON file contains an array of JSON objects in which each object consists of two properties (as shown in Figure 3), namely, "productDetails" and "customerReviews". On the one hand, the property "productDetails" has elements of "Title" and a set of "useful-data parts", according to each EC website. The "Title" element has the name of the product itself. "useful-data parts" have the extracted parts of the EC website, as shown in Table 1. On the other hand, the property "customerReviews" has an array of the review text.

---

[1] https://docs.scrapy.org/en/latest/

```
[
    ..............................................................
    ..............................................................
    {
    'productDetails' :
        {
        'Title' : '2020 HP 14-inch HD Touchscreen Premium Laptop PC, AMD Ryzen 3 3200U Processor, 8GB DDR4 Memory, 256GB SSD, Bluetooth, Windows 10, Silver'
        'About this item': [ 'MD Dual Core Ryzen 3 3200U Processor (2.6GHz, up to 3.5GHz, 4MB cache, 2 cores)', 'GB DDR4 SDRAM, 256GB Solid Sata Drive,
                            AMD Radeon Vega 3 Graphics'],
        'Compare with similar items': [ 'Customer Rating', 'Price', 'Computer Memory Size', 'CPU Speed', 'Display Resolution Maximum', 'Screen Size', 'Hard
                            Disk Size', 'Item Dimensions', 'Operating System', 'Processor Count', 'RAM Type'],
        'Product description': [ 'Designed for long-lasting performance this HP 14-inch laptop lets you speed through tasks and stay connected all day, with the latest
                            processors and a rich HD display. '],
        'Product information': [ 'Screen Size', 'Screen Resolution', 'Hard Drive Interface', 'Power Source', 'Batteries']
        }
    'customerReviews' :
        [
        'Very nice laptop. Arrived quickly and in perfect condition!', 'Very happy with the laptop.', "It is lightweight, has a beautiful and vibrantly colored screen, an easy to
        use keyboard (I work online 15 hours a day, so it can't be too stiff or too difficult to hit the keys), and it is fast. It also literally boots in about 5 seconds and is very
        quiet due to the SSD drive.", 'No complaints at all, and well worth the money spent on it.']
        ]
    },
    ..............................................................
    ..............................................................
]
```

**FIGURE 3.** An example of JSON object from Amazon.

To sum up, our scraper mainly focuses on a semi-supervised process of crawling the EC websites to find product details and reviews from these websites. This step is useful to create our system as the product details are used for extracting the hierarchical aspects (the first phase). Besides, the customer reviews to summarize the hierarchical aspects based on opinions (the second phase). The details of these two phases will be discussed in the next two subsections, respectively.

### 3.3. Hierarchical Aspect Extraction

This section focuses on the Hierarchical Aspect Extraction (HAE) phase of our system. In this phase, popular aspects first are extracted from the product details that are gathered from the EC websites. After that, these aspects are represented in different granularity levels. For example, the "Camera" is an aspect of the "Laptop" domain, whereas "Resolution" and "Lens" are components of "Camera" and not of "Laptop" directly. This phase is implemented in two sequential tasks named (i) aspect extraction and (ii) hierarchical clustering.

**Input:**

$D$: A set of products details of the same product type $\in$ P

$\theta$: Threshold score for aspect clustering

**Output:**

$H$: A hierarchical aspect set

**Method:**

//Task 1. Aspect Extraction

1: Direct aspect set $Ad \leftarrow \phi$, Candidate aspect set $Ac \leftarrow \phi$

2: **for all** product details $d_i \in D$ **do**

3:     $Ad_i \leftarrow$ Parsing $(p_i)$

4:     $Ad \leftarrow Ad \cup Ad_i$

5:     **for all** sentence $s_j \in p_i$ **do**

6:         $T_{i,j} \leftarrow$ POS $(s_j)$

7:         $Ac_{i,j} \leftarrow$ ExtractNouns $(T_{i,j})$

8:         $Ac \leftarrow Ac \cup Ac_{i,j}$

9:     **end for**

10: **end for**

11: $A \leftarrow$ SemanticSimilarity $(Ad, Ac)$

//Task 2. Hierarchical Clustering

12: **let** each aspect $a_i \in A$ is a cluster $c_i \in C$

13: **for all** aspect $a_i \in A$ **do**

14:     **for all** $c_j \in C$ **do**

15:         **if** ClusterSim $(a_i, c_j) > \theta$ **then**

16:             $c_j = c_j \cup a_i$

17:         **end if**

18:     **end for**

19: **end for**

20: update $C$ with distinct clusters

21: **for all** $c_j \in C$ **do**

22:     **for all** aspect $a_i \in c_j$ **do**

23:         **for all** aspect $a_k \in c_j$ **do**

24:             **if** ClusterSim $(a_i, a_k)$ be maximum **then**

25:                 parent aspect is $a_i$ and reminder aspects are children, and update $H$.

26:             **end if**

27:         **end for**

28:     **end for**

29: **end for**

30: **return** $H$

**ALGORITHM 2.** Hierarchical aspect extraction (Phase A).

### 3.3.1. Aspect Extraction

This task fetches the aspects from the product information (i.e., the template of the websites). It has the following advantages. First, templates have attributes that may not be mentioned by the reviewers in their opinions, which avoid problems of not considering important aspects. Second, the templates are provided by manufacturers, so the text does not suffer from problems with spelling, punctuation and grammatical errors, contrary to the reviews' comments. Finally, the manufacturers highlight the most useful product characteristics of their websites.

For extracting product aspects, there are two specific types of them in the product template: direct aspects and indirect aspects. The first type is almost represented in the first column of the <Table> tag from the EC web-source using the proposed method by [33]. On the other hand, the second type is found inside paragraphs and needs to be processed to be extracted.

Algorithm 2 takes products details D of the same type P as input and generates as output the hierarchical aspect set H. This algorithm is represented in two tasks: aspect extraction and hierarchical clustering. For the first task, it is devoted to extracting popular aspects A (Steps 1-11). Step 1 initializes the direct aspect set Ad and candidate aspect set Ac. The algorithm iterates through the product details $d_i \in D$ (Loop 2 – 10), and for each one, the direct aspect $Ad_i$ can be extracted using simple parsing as done by [33] (Step 3) and is added to the Ad set of P (Step 4). The algorithm also iterates through the sentences $s_j \in p_i$ (Loop 5 – 9), in which each sentence $s_j$ produces the part-of-speech tag $T_{i,j}$ for each word using the Stanford log-linear POS Tagger[2] [34] (Step 6). Step 7 selects noun words and noun phrases for tags identified $T_{i,j}$ as candidate aspect $Ac_{i,j}$ and is added to a set of candidate aspects Ac of P (Step 8). Finally, in step 11, the popular aspects A of P are extracted by measuring the similarity between the two sets: candidate aspects Ac and direct aspects Ad, using SemanticSimilarity function. This function implements word embedding due showed great value in handling semantic similarity as in [35], in which each sentence only contains a single aspect or less. The second task will be discussed as follows.

### 3.3.2. Hierarchical Clustering

This task clusters a set of product aspects into a meaningful hierarchy. It applies cluster similarity technique-based word embeddings as in [36], which is encapsulated in the ClusterSim function. This function is the average similarity between each pair of aspects.

Algorithm 2 (Steps 12-30) shows the steps to obtain the hierarchical aspect set H given the popular product aspects A extracted previously (Steps 12-30). Step 12 allows each aspect $a_i \in A$ is a cluster $c_i \in C$. The algorithm iterates through a set of aspects $a_i \in A$ (Loop 13-19) and through a set of clusters $c_j \in C$ (Loop 14-18). For each iteration, if the average similarity between the aspect $a_i$ and the cluster $c_j$ is more than the threshold score θ (Step 15), then $a_i$ is merged with $c_j$ (Step 16), and update C with distinct clusters (Step 20). The algorithm repeats through a set of clusters $c_j \in C$ (Loop 21-29), a set of aspects $a_i \in c_j$ (Loop 22-28) and a set of aspects $a_k \in c_j$ (Loop 23-27). In each iteration, if the average similarity between each pair of aspects ($a_i$, $a_k$) is the maximum (Step 24), then $a_i$ is a parent aspect. Otherwise, the others are children and finally update the hierarchy H (Step 25) to be returned in Step 30.

*3.4. Hierarchical Aspect-based Opinion Summarization*

In SEOpinion, the second phase is introduced to provide a hierarchical-aspect set *H* extracted from the previous phase with a set of opinionated sentences $O_i$, where each sentence is classified according to the polarity of its aspects associated in *H*. This phase is organized into three sequential subtasks, named (i) subjectivity detection, (ii) opinion mapping and (iii) aspect-level polarity detection. These tasks are shown in Algorithm 3, which takes the hierarchical aspect set *H* and a set of reviews $R_i$ for each product $p_i$ as input and generates a hierarchical aspect-based opinion summary $S_i$, which is also repeated on all products in *P*.

### 3.4.1. Subjectivity Detection

This task is often called an opinion extraction, which is used to differentiate sentences that state opinions from those that state facts. To address this task, our approach assumes that an opinion sentence $O_i$ in a review $R_i$ must include at least one noun and one adjective. Consequently, it includes three steps: part-of-speech (POS) tagging, sentence filtering and word-based semantic score.

---

[2] The tool can be found at https://nlp.stanford.edu/software/tagger.shtml

**Input:**

*H:* A hierarchical aspect set

$R_i$: A set of reviews from a given product $p_i \in P$

$\theta$: Threshold score

**Output**:

$S_i$: A hierarchical aspect-based summary $\in p_i$

**Method:**

//Task 1. Subjectivity Classification

1: opinion sentences set $O_i \leftarrow \phi$, mapped aspect set $M_i \leftarrow \phi$, hierarchical aspect-based summary $S_i \leftarrow \phi$

2: **for each** review $r \in R_i$ **do**

3:     $S_T \leftarrow \{sentence\ s\ |\ s \in r \wedge POS\ (s)\}$

4:     **if** $S_T$ has NN and ADJ **then**

5:         $P' \leftarrow 0$

6:         $N' \leftarrow 0$

7:         **for all** word $w \in S_T$ **do**

8:             $P, N \leftarrow SemanticScore\ (w)$

9:             $P' \leftarrow P' + P$

10:            $N' \leftarrow N' + N$

11:        **end for**

12:        **if** $P'$ or $N' > \theta$ **then**

13:            $O_i \leftarrow O_i \cup S_T$

14:        **end if**

15:    **end if**

16: **end for**

//Task 2. Aspect Mapping

17: $I_i^+ \leftarrow \phi$

18: **for all** parent aspect $Ap_j \in H$ **do**

19:    **let** opinion sentence $o \in O_i$

20:    **if** *mapped ($Ap_j$, o) $\leftarrow$ true* **then**

21:        **for all** child aspect $Ac_k \in Ap_j$ **do**

22:            **if** *mapped ($Ac_k$, o) $\leftarrow$ true* **then**

23:                $M_i \leftarrow M_i \cup <Ac_k, o>$

24:            **end if**

25:        **end for**

26:    **end if**

27: **end for**

//Task 3. Aspect-level Sentiment Classification

28: **for all** $m_j \in M_i$ **do**

29:    $P_j \leftarrow classifyPolarity\ (m_j)$

30:    $S_i \leftarrow S_i \cup P_j$

31: **end for**

32: **return** $S_i$

**ALGORITHM 3.** Hierarchical aspect-based opinion summarization (phase B).

As shown in Algorithm 3, this task extracts a set of opinionated sentences $O_i$ from reviews $R_i$ of a specific product $p_i \in P$. The algorithm iterates through review $r \in R_i$ (Loop 2-16). In Step 3, function *POS* determines part-of-speech tagging for each sentence $s \in r$, using the Stanford Log-linear POS Tagger [34]. The set of these tagged sentences is designed as $S_T$. Step 4 checks if the tagged sentence $S_T$ has at least one noun *NN* and one adjective *ADJ*, then a semantic score gives a measure of the number of objective and subjective words in the sentence (Loop 7-11), which is encapsulated in the *SemanticScore* function (Step 8). This function is adopted using SENTIWORDNET [37], where each word contains positive (*P*) and negative (*N*) scores for each one to determine subjective words in the sentence. The positive and negative scores are summed over all noun and adjective words in the sentence (Steps 9-10) and used to normalize the individual scores of each word. The normalized scores are compared to a threshold $\theta$. If the word has a positive/negative score larger than this threshold (Step 12), then the sentence is defined as an opinionated sentence. These steps are performed on all sentences of each review to construct a full set of opinion sentences $O_i$ for the specific product $p_i$ (Step 13).

3.4.2. Aspect/Opinion Mapping

This task maps opinionated sentences according to the specific aspects that have been given by manufacturers. Hence, there are two cases for this task in our system: i) the opinion may indicate the aspect category as a whole (General) or ii) one of the aspect terms (child aspects) of the aspect category. Proposals in [23] and [38], demonstrated the mapping of opinions in one level aspects. In contrast, our approach maps the opinionated sentences $O_i$ according to their aspects in two-levels of *H*.

As shown in Algorithm 3, this task receives a hierarchical aspect set *H* and opinionated sentences $O_i = \{o_1, o_2 \ldots o_n\}$, which were extracted from the previous, and return a mapped aspect set $M_i$ (Steps 17-27). The algorithm iterates through the set of parent aspects $A_p$ inside the hierarchical aspect set *H* (Loop 18-27). In Step 20, if a parent aspect $A_{p_j}$ is mapped to an opinionated sentence *o* as applied in [38], the algorithm also repeats the process with the child aspects $A_c$ associated with its parent aspect (Loop 21-25). In each iteration, if a child aspect $A_{c_k}$ is mapped to an opinion sentence *o* (Step 22), it stores in a mapping aspect set $M_i$ in pairs of the child aspect $A_{c_k}$ and its opinion *o* (Step 23).

3.4.3. Aspect-Level Polarity Detection

The main subtasks of sentiment classification are emotion identification [39], predicting sentiment intensity [40], and polarity detection [41]. Emotion identification detects the emotions behind sentiments such as anger or sadness. Predicting sentiment intensity seeks to identify the polarity degree (e.g., 'good', 'wonderful', and 'awesome'). Furthermore, polarity detection classifies text as negative or positive. Hence, our system applies the polarity detection task on the aspect, which classifies polarity for each opinionated sentence based on the aspect that matches it. Polarity detection has been solved using different machine learning techniques (unsupervised [42], semi-supervised [10], [43] and supervised [44]). Recently, Deep Learning (DL) techniques have achieved success in polarity detection [45], especially with the use of BERT embedding [46]. Therefore, our work uses DL-based BERT representation.

Algorithm 3 receives a map aspect set $M_i$ from the previous task, and returns a hierarchical aspect-based opinion summary $S_i$ (Steps 28-32). The algorithm iterates through the set of the mapped aspect set $m_i$. (Loop 28-31). For each iteration, in step 29, the task classifies the polarity of opinion (positive or negative) according to its child aspect as in [46] using *classifyPolarity* function. In step 30, the polarity $P_i$ extracted from this function is added to the summary $S_i$. Finally, it generates the hierarchical aspect-based summary $S_i \in p_i$, which contains a set of aspect terms associated with their classified sentences (Step 32), as shown in Figure 1(d).

*3.5. User Interface*

Our system introduces a comparison among a set of products of the same type by providing a summarization and an exploration interface, as shown in Figure 4. This interface enables the user to browse through several product aspects and go through related opinions. The figure displays a screenshot of the SEOpinion interface showing information related to "Laptop". The interface consists of three panels, namely (i) the product presentation panel; (ii) the aspect-opinion-summarization panel; and (iii) the sentence-opinion-exploration panel.

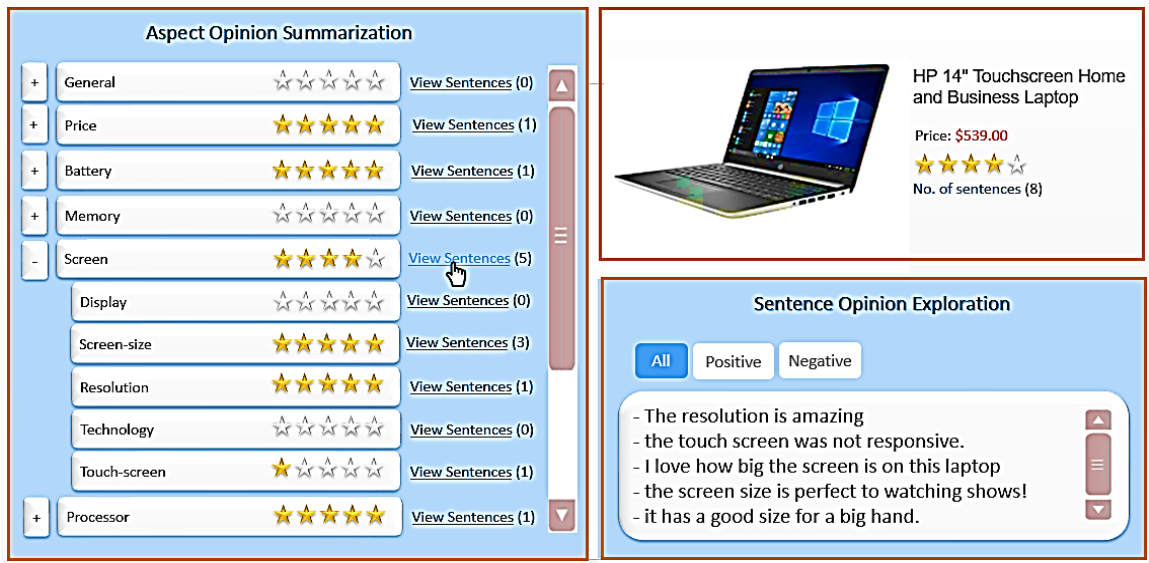

**FIGURE 4.** Screenshot of our SEOpinion system.

- The product presentation panel shows information about the product, such as its name, price, images, rate summary of its opinion sentences, and the number of these sentences in the top-level aspects (i.e., General, Price, Battery, Memory, Screen, and Processor).
- The summarization panel displays the hierarchy aspects of the product and the aspect-based summary. Initially, sub-aspects are kept hidden until the user clicks on the related parent aspect. For example, "*Display*", "*Screen-size*"; "*Resolution*"; "*Technology*"; and "*Touch-screen*" are components or sub-aspects of the "*Screen*". For each top-level aspect, the total of sentences on each aspect is shown because it gives other users the confidence of the aspect rate (i.e., when the number of sentences increased, the user's confidence in the rating aspect increased). Furthermore, the rated summary for each aspect is the average for the scores of its sentences. Our system considers positive = 5 and negative = 1. For example, as shown in Figure 4, the aspect "*Screen*" contains five sub-aspects for 5 sentences, which include four positive and one negative. Thus, the average of all sentences for summarizing the aspect "*Screen*" is calculated as (5 + 5 + 5 + 5 + 1) / 5 = 4.2.
- The exploration panel shows the opinionated sentences that are categorized as positive or negative. Initially, this panel does not display these sentences of the product if no product aspect is selected. These sentences related to the aspect are shown in this panel only when the user clicks on the "view sentences" button of the aspect in the summarization panel.

This way, the user is not distracted with irrelevant aspects. The valued-added feature of this interface is that it allows the interaction between users and content, which makes it easy for the end-user to identify product aspects of interest and focus on sentences that contain relevant aspect information.

## 4. Experiments

This section describes several experiments that have been conducted to evaluate the performance of our SEOpinion system. First, the datasets and the preprocessing steps are presented, respectively. Second, the baseline methods are introduced. Third, the evaluation metrics are described for the proposed approach, and finally, the experiment setups are depicted.

### 4.1. Data Collection and Preprocessing

A collection of products/items from the same product type (e.g., Book or Camera) on the EC websites is needed to show the effectiveness of the proposed approach. Each website contains product details identified with aspect categories and aspect terms (as shown in Figure 1(b)), and review sentences are labeled with aspects and polarities/sentiments (as shown in Figure 1(d)). As there is no such benchmark corpus, we create a Laptop Collection from five EC websites (LC5) dataset crawled from *Amazon, Flipkart, eBay, Walmart*, and *BestBuy* website, which will make it publicly available for research. On each item, the aspect terms are manually fetched from product details and the sentences of reviews are annotated to the extracted aspects. Each sentence belonging to one aspect is also labeled as expressing positive or negative sentiment (ignoring the scores of neutral, since it is not useful for the aspect summarization). Details of the dataset are shown in Table 2, which shows the distribution of aspect terms, aspect categories and sentences with polarity for each one product item.

Customer reviews are usually full of noise, unstructured, spelling errors, arbitrariness, short text, and incomplete grammatical structures due to the frequent presence of irregular grammar, malformed words, acronyms, and non-dictionary terms, etc. These factors are the most common problem in customer reviews and will affect the performance of sentiment analysis tasks. Therefore, a series of preprocessing on our LC5 dataset is performed by removing all numbers, stop words, all non-ASCII and English characters, and all URL links. Followed by replacing negative references and emoticons, then by slang to their full word form and expanding acronyms. Finally, the Natural Language Toolkit (NLTK) [47] was adopted tokenization. After being preprocessed, the LC5 dataset is ready for testing our approach.

**TABLE 2.** Statistics of our LC5 dataset.

| Dataset (Domain) | No. Laptop reviewed Items (web pages) | For each one item (Average) | | | | Polarity | |
|---|---|---|---|---|---|---|---|
| | | No. Aspect terms | No. Aspect Categories | No. Sentences | No. Sentences/ aspect term | Positive % | Negative % |
| Amazon | 707 | 42 | 11 | 3,289 | 77.3 | 62 | 38 |
| Flipkart | 284 | 55 | 8 | 546 | 6.9 | 69 | 31 |
| eBay | 856 | 74 | 3 | 11 | 0.12 | 73 | 27 |
| Walmart | 790 | 18 | 6 | 2,180 | 97.1 | 62 | 38 |
| BestBuy | 525 | 72 | 17 | 3,574 | 43.4 | 67 | 33 |

*4.2. Baseline*

The experiments have been conducted using three state-of-the-art baseline classifiers: Support Vector Machine (SVM), Convolutional Neural Network (CNN), and Recurrent Neural Network (RNN). The SVM classifier is a state-of-the-art traditional machine learning method, which exploits input features such as uni/bigram features and part-of-speech (POS) tags, as in [48]. On the other hand, CNN and RNN utilized the word embedding as an input feature, in which the embedding was trained using random initialization, the GloVe [49] and the BERT embedding [5]. The GloVe was used in our work, which achieved better results than Word2Vec [45]. The pre-train BERT word embedding [5] had been used on the Amazon corpus.

*4.3. Evaluation Measures*

The baseline methods are evaluated on the two main phases of our system: (i) hierarchical aspect extraction and (ii) hierarchical aspect-based opinion summarization. In both, the performance is measured using the metrics of recall (R), precision (P), and F1-measure (F). These metrics are calculated through the confusion matrix, as shown in Table 3. The table shows a confusion matrix for two classes (Positive, Negative), where TP (True Positive) and TN (True Negative) are sampled correctly. In contradiction, FP (False Positive) and FN (False Negative) are incorrect. These metrics are shown in Eqs. (1) (2) and (3) that are commonly used for sentiment analysis performance. Recall measures the percentage of labels found by the system. Precision measures the percentage of labels correctly, assigned by the system. F1-measure is based on precision and recall for presenting the right results. From another perspective, the baselines are evaluated on each task in the two main phases through an accuracy metric. Accuracy represents the correct results, as shown in Eq. (4). The results of the baselines are the average of 10 test scores.

$$\text{Recall} = \frac{TP}{TP+FN} \tag{1}$$

$$\text{Precision} = \frac{TP}{TP+FP} \tag{2}$$

$$\text{F1-measure} = \frac{2 * Precision * Recall}{Precision + Recall} \tag{3}$$

$$\text{Accuracy} = \frac{TP+TN}{TP+TN+FP+FN} \quad (4)$$

**TABLE 3.** Confusion matrix for two classes.

|  |  | Prediction Label | |
|---|---|---|---|
|  |  | *Positive* | *Negative* |
| **Actual Label** | *Positive* | TP | FN |
|  | *Negative* | FP | TN |

*4.4. Experiment Setups*

Our experiments have tuned the settings of the CNN and RNN models, where these models were implemented in PyTorch [50] framework[3], and on a single NVIDIA Tesla P100 GPU's. Besides, the two models were applied to two types of embedding layer: GloVe and BERT. For the learning process, the GloVe embedding has been used as in [51]. On the contrary, the BERT embedding has been fine-tuned, in which the dropout probability was kept at 0.1 [5] . Besides, Adam optimizer was used to update the model parameters [51]. The best hyper-parameter batch size and learning rate were obtained from {16, 32} and {2e-5, 3e-5} respectively by grid search. Finally, the details of used hyper-parameters and the configurations are shown in Table 4. For each experiment, ten-fold cross-validation was applied 100 times for each website of our dataset. Besides, the average accuracy was obtained by observing 100 replications of cross-validation. An imbalance in our LC5 dataset is the main problem, in which positive reviews not equal to negative reviews. To address this problem, we performed a sampling of various sub-datasets and took an average of outcomes for each one.

**TABLE 4.** The details of used hyper-parameters and the configurations on deep learning-based BERT methods.

| | |
|---|---|
| *Word Embedding* | *BERT* [5] |
| *Dropout Rate* | *0.1* |
| *Batch Size* | *Search from = {16, 32}* |
| *Learning Rate* | *Search from = {2e-5, 3e-5}* |
| *Max Epoch* | *6* |
| *Max Sequence Length* | *128* |
| *Optimizer* | *Adam* [53] |
| *Embedding Layer Dimension* | *768* |
| *Deep Learning Framework* | *Pytorch* [52] |

## 5. Experimental Results and Analysis

The objective or our analysis is to comprehensively evaluate the performance of the SEOpinion system in the above two sequential phases: hierarchical aspect extraction (HAE) and hierarchical aspect-based opinion summarization (HAOS). The results were analyzed and achieved by SEOpinion when applied to our LC5 dataset. The results of both of them are displayed in Table 5 and Table 6, respectively. The highest performance scores are underlined in the two tables. These results are achieved on our LC5 dataset and are shown in three parts: Part I, SVM is based on the traditional feature-engineering method [52]. Part II and part III contain deep learning methods (CNN and RNN) based on word embedding, including pre-trained word vectors (Random initialization, GloVe [49] and BERT [5]).

*5.1. Results for Hierarchical Aspect Extraction*

This stage uses two sequential tasks for creating a hierarchical aspect set, such as aspect extraction and hierarchical aspect clustering. Table 5 shows our results for the HAE phase on the LC5 dataset. As it can be seen from the table, the SVM method is the worst performance for the F-measure of the five product websites. In contrast, the RNN method with BERT embedding is the best, in which the highest F1-measure is 80.9% in the Amazon, and the minimal F1-measure is 73.3% in eBay. Moreover, the F1-measure of the RNN+BERT outperforms SVM, CNN, CNN+GloVe, RNN, and RNN+GloVe on the Amazon, eBay, Walmart, and BestBuy type. The RNN+BERT is close to RNN+GloVe on all websites of our dataset.

---

[3]  https://pytorch.org/

**TABLE 5.** Comparison results for Hierarchical Aspect Extraction phase in our system on our LC5 dataset.

| Model | Text Representation | Amazon | | | Flipkart | | | eBay | | | Walmart | | | BestBuy | | | Avg. |
|---|---|---|---|---|---|---|---|---|---|---|---|---|---|---|---|---|---|
| | | P | R | F | P | R | F | P | R | F | P | R | F | P | R | F | |
| SVM | Hand-crafted Features [50] | 62.4 | 62.5 | 62.4 | 64.0 | 65.0 | 64.5 | 62.4 | 62.5 | 62.4 | 62.3 | 62.1 | 62.2 | 59.4 | 59.5 | 59.4 | 62.2 |
| CNN | Random (Embedding) | 72.4 | 66.0 | 69.1 | 70.3 | 71.0 | 70.6 | 63.2 | 62.6 | 62.9 | 60.5 | 58.9 | 59.7 | 57.1 | 53.0 | 55.0 | 63.5 |
| CNN | GloVe [55] (Embedding) | 73.1 | 66.8 | 69.8 | 71.9 | 79.8 | 75.6 | 64.4 | 63.9 | 64.1 | 70.9 | 64.0 | 67.3 | 68.5 | 62.5 | 65.4 | 68.5 |
| CNN | BERT (our) (Embedding) | 79.6 | 73.3 | 76.3 | 72.1 | 79.8 | 75.8 | 72.5 | 68.8 | 70.6 | 72.9 | **79.9** | 76.2 | 72.8 | 75.8 | 74.3 | 74.7 |
| RNN | Random (Embedding) | 72.7 | 72.8 | 72.7 | 74.8 | 75.0 | 74.9 | 73.3 | 73.2 | 73.2 | 71.6 | 73.9 | 72.7 | **76.1** | 72.1 | 74.0 | 73.5 |
| RNN | GloVe [55] (Embedding) | 82.4 | 76.0 | 79.1 | **80.3** | 81.0 | 80.6 | 73.2 | 72.6 | 72.9 | 70.5 | 68.9 | 69.7 | 67.1 | 63.0 | 65.0 | 73.5 |
| RNN | BERT (our) (Embedding) | **83.3** | **78.7** | **80.9** | 77.7 | **84.1** | **80.8** | **73.4** | **73.3** | **73.3** | **77.5** | 75.7 | **76.6** | 74.4 | **75.9** | **75.1** | **77.4** |

*5.2. Results for Hierarchical Aspect-Based Opinion Summarization*

This stage uses the three sequential tasks for summarizing the extracted hierarchical aspect set by opinions, such as subjectivity classification, opinion mapping and aspect-level polarity detection. Table 6 displays our results for the HAOS phase on our LC5 dataset. As it can be seen from the table, the SVM method is the worst performance for the F1-measure on all EC websites because of the SVM method using feature extraction of sentiment analysis tasks. On the other hand, the RNN method with BERT embedding is the best, in which the highest F1-measure is 86.0% in the Amazon, and the minimal F1-measure is 78.7% in Flipkart. Moreover, the F1-measure of the RNN+BERT outperforms SVM, CNN, CNN+GloVe, RNN, and RNN+GloVe on four websites out of five. The RNN+BERT is close to RNN+GloVe on the Flipkart website.

**TABLE 6.** Comparison results for Hierarchical Aspect-based Opinion Summarization phase in our system on our LC5 dataset.

| Model | Text Representation | Amazon | | | Flipkart | | | eBay | | | Walmart | | | BestBuy | | | Avg. |
|---|---|---|---|---|---|---|---|---|---|---|---|---|---|---|---|---|---|
| | | P | R | F | P | R | F | P | R | F | P | R | F | P | R | F | |
| SVM | Hand-crafted Features [50] | 73.9 | 74.1 | 74.0 | 72.3 | 75.3 | 73.8 | 65.6 | 61.0 | 63.2 | 71.5 | 73.5 | 72.5 | 71.4 | 73.4 | 72.4 | 71.2 |
| CNN | Random (Embedding) | 79.2 | 89.7 | 84.1 | 79.3 | 75.7 | 77.5 | 64.9 | 55.5 | 59.8 | 76.5 | 76.0 | 76.2 | 74.6 | 79.2 | 76.8 | 75.0 |
| CNN | GloVe [55] (Embedding) | 82.6 | 78.3 | 80.4 | 80.3 | 78.7 | 79.5 | 73.2 | 72.6 | 72.9 | 77.1 | 63.0 | 69.3 | 79.5 | 84.1 | 81.7 | 76.9 |
| CNN | BERT (our) (Embedding) | 83.1 | **88.8** | 85.9 | 80.9 | 74.0 | 77.3 | 78.3 | 77.0 | 77.6 | 78.5 | 72.5 | 75.4 | 81.2 | 83.4 | 82.3 | 79.7 |
| RNN | Random (Embedding) | 83.1 | 76.8 | 79.8 | 81.0 | 75.3 | 78.0 | 74.4 | 73.9 | 74.1 | **80.9** | 74.0 | 77.3 | 78.5 | 72.5 | 75.4 | 77.0 |
| RNN | GloVe [55] (Embedding) | 84.5 | 81.7 | 83.1 | **81.6** | 78.3 | **79.9** | 82.5 | 78.8 | 80.6 | 78.7 | 75.5 | 77.1 | 80.5 | 75.7 | 78.0 | 79.8 |
| RNN | BERT (our) (Embedding) | **86.1** | 85.9 | **86.0** | 76.3 | **79.8** | 78.0 | **84.5** | **85.3** | **84.9** | 80.0 | **77.5** | **78.7** | **82.0** | **88.5** | **85.1** | **82.6** |

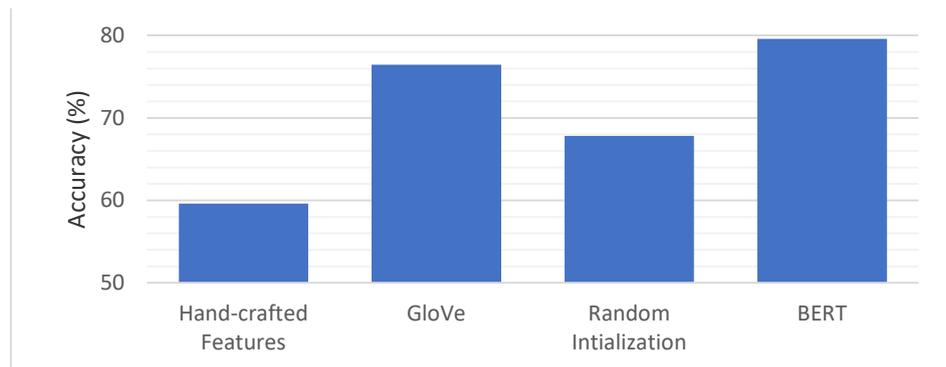

**FIGURE 5.** Accuracies from the four representation types.

*5.3. Analysis of Results*

From another perspective, through an accuracy metric, we study which text representations achieved the best results. The impact of the four different representations is investigated on our proposed approach, namely, the Hand-crafted Features [52], GloVe embeddings [49], Random Initialization embeddings, and BERT embeddings [5]. Figure 5 shows the performance of all four different representations on our LC5 dataset. The GloVe embeddings and BERT embeddings based deep learning models (i.e., RNN and CNN) are studied and achieved the best results of the previous figure. Therefore, the comparison of the two word embeddings-based DL models (i.e., CNN and RNN) on both phases shows the average performance for each product website, as shown in Figure 6. As can be seen, the RNN-based BERT method has the highest accuracy and the CNN based-GloVe method has the lowest accuracy among all websites. More generally, on both RNN and CNN, the BERT method has increased the accuracy of the GloVe method on average by at least 3.1%, approximately. Also, the accuracy difference between GloVe embedding and BERT embedding increased when the volume of data has grown, such as BestBuy, Amazon, and Walmart websites. To study in-depth the size of the data, Figure 7 shows the overall picture of BERT embedding and GloVe embedding on the average of both phases. Note that the BERT embedding achieved the best results for large data sizes. In contrast, GloVe embedding had the best accuracy rates obtained for small data volumes. For more generally, when studying the baseline models' performance on the five tasks of our system, separately, whose names are aspect extraction, hierarchical clustering, subjectivity detection, opinion mapping, and aspect level-polarity detection. As shown in Figure 8, BERT embedding-based two DL models provided a significant increase in accuracy metric over GloVe embedding, while noticeably superior to other baselines on different tasks. Also, the aspect extraction task and hierarchical clustering task are the least accurate comparison of others due to little retrieved information from product template websites. Thus, our intention in the future is to try the proposed approach on another domain (movies or restaurants).

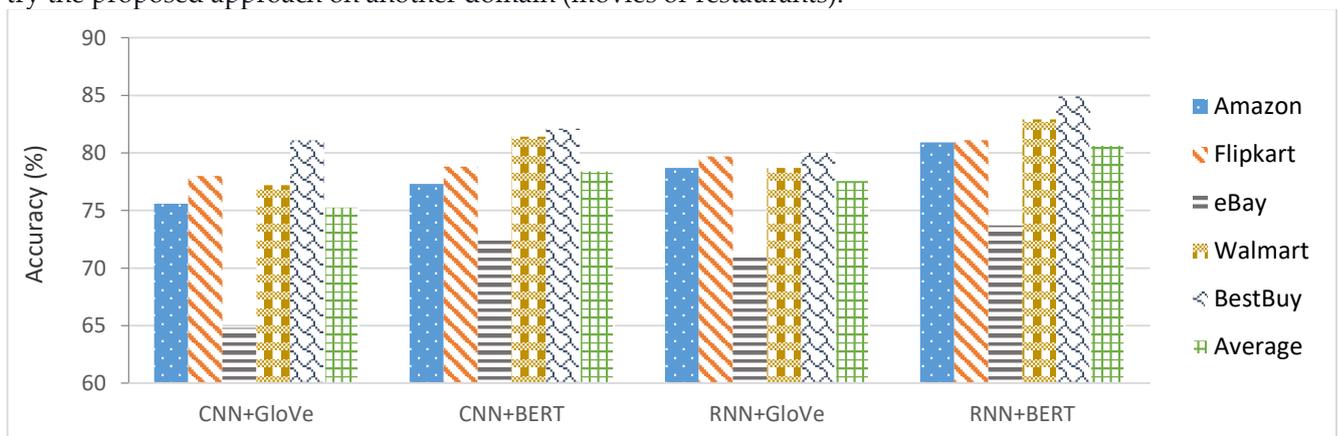

**F IGURE 6.** Comparisons of the two word embeddings-based deep learning models for our LC5 dataset.

To sum up, from the presented results, we can observe that DL techniques are more flexible than the SVM approach with hand-crafted features. Furthermore, the BERT embedding showed significance in performance DL models, especially the RNN model. Besides, BERT can benefit from a larger training set compared to GloVe.

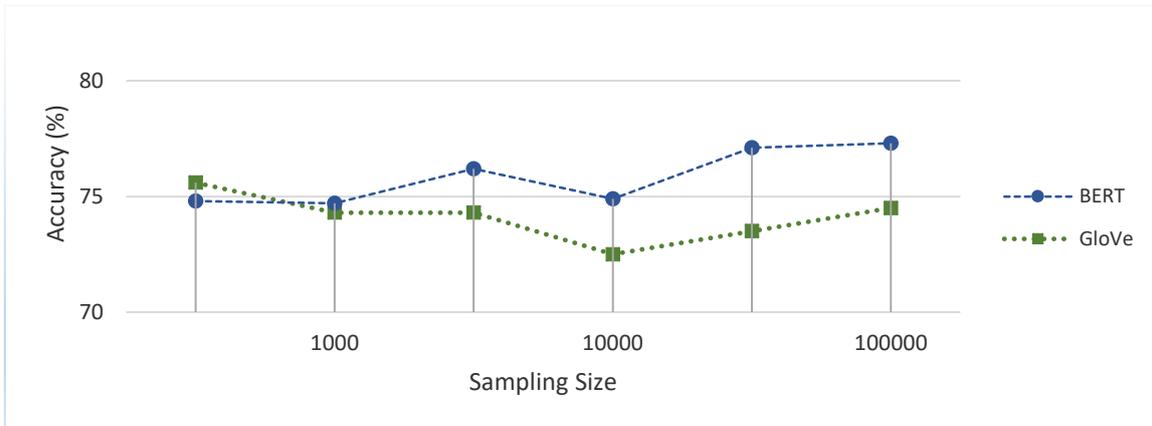

**FIGURE 7.** Accuracies from sampling of different size for our dataset.

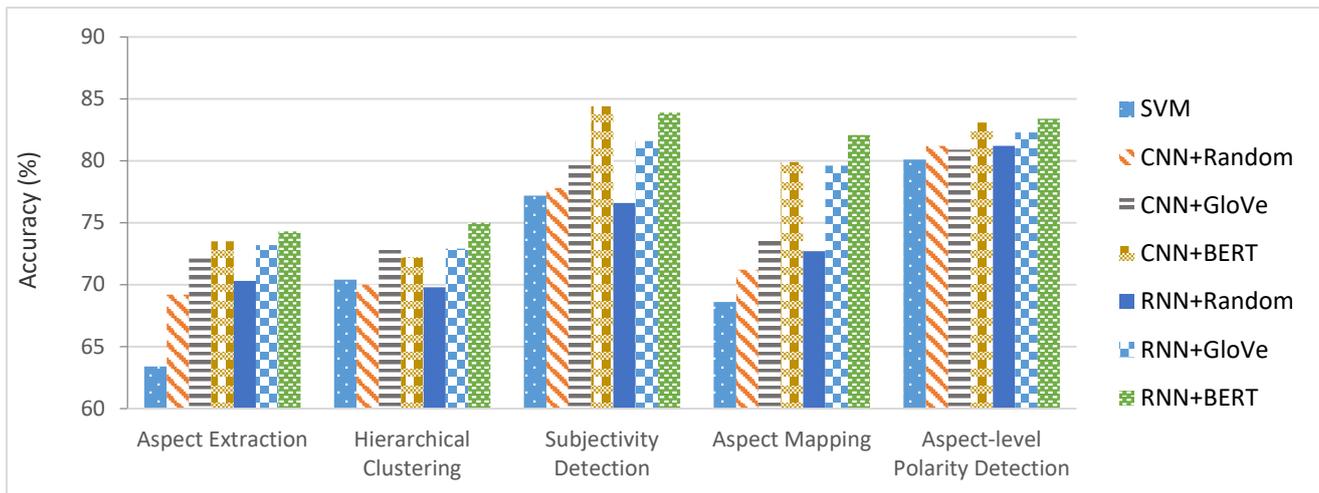

**FIGURE 8.** Comparisons of the baseline models for the five research tasks.

## 6. Limitations and Future Directions

The conducted experiments achieved the best results using the RNN-based BERT approach. However, our results were not the best among the existing summarization systems [16], [20] because they ignored the multi-granularity of aspects and used pre-defined aspects. Moreover, our approach has several limitations that could be summarized as follows:

- The system only works well with the web pages that have much of product details (aspects) embedded in the page' templates. Therefore, the little result was on the eBay website, as shown in Figure 6, which has less details in its webpage templates than others.
- Implicit aspects are difficult to be extracted from the template, unlike reviews. In [53] a rule-based approach to obtain both explicit and implicit aspects from customer reviews was proposed.
- The opinion mapping task in our system worked on matching an opinion sentence with only one aspect. Some opinion sentences may express more than one aspect. For example, the opinion sentence "my phone is good for its price and performance" is associated with two aspects "price" and "performance".

For future work, we plan to investigate strategies for improving the performance of our system, as follows: (i) Using multichannel embedding [54] on various DL models as RNN and CNN, where the pre-trained word embeddings are directly incorporated into the word embedding matrix. The advantages of the multichannel embedding are that it can provide rich semantic/sentiment representations and avoid word embedding interference. Thus, our expectation is that multichannel embedding will give better results than single embedding. (ii) The aspect extraction task may be improved by incorporating aspects obtained from the template with the other aspects (explicitly or implicitly) mentioned in the customer's reviews.

**7. Conclusion**

This paper investigated the effectiveness of the BERT embedding component with the CNN model and the RNN model for creating our Summarization and Exploration Opinion (SEOpinion) system. SEOpinion compares a set of products in EC websites from the same type on two phases: (i) Hierarchical Aspect Extraction (HAE) and (ii) Hierarchical Aspect-based Opinion Summarization (HAOS). Hence, the experimental results demonstrated the superiority of the BERT embedding-based RNN and CNN model in both phases, where it is better than GloVe embedding by up to 3.1% on our LC5 dataset. Our system only focused on comparing a group of products from the same type in terms of summarizing the aspect's opinions. In the future, we will plan to apply it to other domains (as movies or restaurants). The proposed system is expected to work well with these domains.

**Acknowledgment**

The authors would like to thank the European Regional Development Fund (ERDF) and the Galician Regional Government, under the agreement for funding the Atlantic Research Center for Information and Communication Technologies (AtlantTIC), and the Spanish Ministry of Economy and Competitiveness, under the National Science Program (TEC2017-84197-C4-2-R).